\documentclass[sigconf]{acmart}

\setcopyright{acmlicensed}
\copyrightyear{2027}
\acmYear{2027}
\acmConference[KDD '27]{KDD 2027}{August 2027}{San Jose, CA, USA}
\acmBooktitle{KDD '27: KDD 2027, August 2027, San Jose, CA, USA}
\acmDOI{XXXX}
\acmISBN{XXXX}
\usepackage{subfig}
\usepackage{mathtools}
\usepackage{bm}
\usepackage{colortbl}
\usepackage{multirow}
\usepackage{pifont}
\usepackage{placeins}
\usepackage{tabularx}

\definecolor{Gray}{gray}{0.9}
\newcommand{\cmark}{\ding{51}}
\newcommand{\xmark}{\ding{55}}
\newcommand{\abcell}[1]{\makebox[1cm][c]{#1}}
\usepackage{amsthm}
\usepackage{enumitem}
\theoremstyle{definition}
\newtheorem{assumption}{Assumption}

\usepackage[ruled,vlined]{algorithm2e}
\AtBeginDocument{%
  \exhyphenpenalty=0
}

\title{SpeCal: Spectral Calibration for Unsupervised RLVR}

\author{Kai Tang}
\email{kai.t@zju.edu.cn}
\affiliation{%
  \institution{Zhejiang University, Ant Group}
  \city{Hangzhou}
  \state{Zhejiang}
  \country{China}
}

\author{Xing Zheng}
\affiliation{%
  \institution{Ant Group}
  \city{Hangzhou}
  \state{Zhejiang}
  \country{China}
}

\author{Haobo Wang}
\authornote{Corresponding author.}
\email{wanghaobo@zju.edu.cn}
\affiliation{%
  \institution{Zhejiang University}
  \city{Hangzhou}
  \state{Zhejiang}
  \country{China}
}

\author{Guangcheng Zhu}
\affiliation{%
  \institution{Zhejiang University, Ant Group}
  \city{Hangzhou}
  \state{Zhejiang}
  \country{China}
}

\author{Shenzhi Yang}
\affiliation{%
  \institution{Zhejiang University, Ant Group}
  \city{Hangzhou}
  \state{Zhejiang}
  \country{China}
}

\author{Zhengqing Zang}
\affiliation{%
  \institution{Zhejiang University, Ant Group}
  \city{Hangzhou}
  \state{Zhejiang}
  \country{China}
}

\author{Yingfan Ma}
\affiliation{%
  \institution{Ant Group}
  \city{Hangzhou}
  \state{Zhejiang}
  \country{China}
}

\author{Cheng Peng}
\affiliation{%
  \institution{Zhejiang University}
  \city{Hangzhou}
  \state{Zhejiang}
  \country{China}
}

\author{Bowen Song}
\affiliation{%
  \institution{Ant Group}
  \city{Hangzhou}
  \state{Zhejiang}
  \country{China}
}

\author{Weiqiang Wang}
\affiliation{%
  \institution{Ant Group}
  \city{Hangzhou}
  \state{Zhejiang}
  \country{China}
}

\renewcommand{\shortauthors}{Tang et al.}

\begin{document}
\pagestyle{plain}
\begin{abstract}
Reinforcement learning with verifiable rewards (RLVR) has substantially improved the reasoning abilities of large language models, but its dependence on ground-truth answers or external verifiers limits its ability to exploit abundant unlabeled reasoning data. Unsupervised RLVR alleviates this bottleneck by deriving scalar rewards from the model's own rollouts. However, the reliability of such self-generated supervision varies substantially across rollout groups and individual responses, while existing methods construct or refine their rewards mainly from decoded outcomes and reward statistics. These endpoint signals reveal what answers the model produces, but provide limited information about whether the underlying rollout representations form a reliable structure for learning.
Our key observation is that hidden-state spectra offer a complementary view of supervision reliability by characterizing both the organization of a rollout group and the support received by each rollout within that group. Based on this observation, we propose \textbf{SpeCal}, a bi-level spectral calibration framework for unsupervised RLVR. At the group level, SpeCal combines spectral mass with a Gaussian band-pass gate over normalized entropy-based effective rank, favoring rollout groups with substantial representation variation whose effective dimensionality lies in a preferred regime. At the rollout level, the same group effective rank determines how many dominant directions are retained in a leave-one-out spectral subspace formed by the remaining rollouts, within which the conditional support of each held-out rollout is measured. The resulting signals calibrate the original reward-induced advantages at the group and rollout levels without redefining the underlying unlabeled reward.
Extensive experiments across multiple reasoning benchmarks and LLM backbones demonstrate that SpeCal consistently outperforms the strongest unsupervised RLVR baselines, yielding average gains of $2.7$ points in-domain and $1.1$ points out-of-domain, while matching or surpassing fully supervised methods in selected settings.
\end{abstract}


\maketitle

\section{Introduction}

Reinforcement learning with verifiable rewards (RLVR) has become an effective paradigm for improving the reasoning ability of large language models (LLMs)~\cite{gpto1,deepseekr1,gemini2.5,qwen3}, especially in domains where final outcomes can be checked by rule-based verifiers or executable tests~\cite{wen2025reinforcement,deepseekmath}. However, the scalability of standard RLVR is constrained by the availability of externally verifiable training questions, including ground-truth answers, executable tests, and reliable domain-specific verifiers~\cite{su2025expanding,peng2025rewarding,MoPPS}. This requirement limits its ability to exploit the large amount of unlabeled reasoning data. Unsupervised RLVR aims to relax this bottleneck by deriving training signals from the model's own rollouts on unlabeled questions, preserving the reward-based optimization paradigm of RLVR while reducing the dependence on external supervision.

Most unsupervised RLVR methods derive pseudo-rewards from output-derived signals computed over a group of sampled rollouts, including majority voting~\cite{ttrl}, predictive confidence~\cite{entropy_base}, and inter-response agreement~\cite{coreward}. Despite their different reward constructions, these approaches generally aggregate output evidence using predefined rules, without explicitly modeling how supervision reliability varies at different granularities. Yet the reliability of self-generated supervision can differ substantially across questions, answer candidates, and individual rollouts.
To this end, recent methods refine these signals at multiple granularities using richer statistics derived from decoded outputs. DARE retains the full empirical answer distribution rather than reducing it to a single majority outcome, while RESTRAIN modulates candidate- and question-level contributions using answer frequency and group consistency~\cite{dare,yu2026restrain}. Although these refinements preserve more output-side information, their reliability estimates remain functions of compressed summaries of decoded outcomes or predictive confidence. However, compressing a rollout group into these output-derived summaries is inherently information-losing: two groups can have identical final-answer histograms and confidence statistics while differing substantially in their internal generation processes. Once this information has been discarded by the compression, no downstream estimator that observes only the resulting summaries can recover it when assessing the reliability of self-generated supervision.

Fortunately, the model's hidden states offer a complementary view of these differences. Studies of individual hidden states have shown that internal states distinguish truthful from hallucinated generations~\cite{azaria2023internal,duan2024hallucination,inside}, while Chain-of-Embedding characterizes how hidden states evolve across layers and uses the resulting latent trajectory for label-free response evaluation~\cite{wang2025coe}. Together, these studies demonstrate that hidden states encode rich signals about the model's internal generation dynamics and the reliability of its responses. However, they predominantly evaluate responses in isolation, whereas unsupervised RLVR needs to derive supervision through relative comparisons among multiple rollouts. The relevant object is therefore not an isolated hidden state, but the joint geometry formed by the rollout representations. For a set of centered representations, the eigenspectrum of their Gram matrix provides a compact and permutation-invariant description of this geometry: spectral mass measures the total inter-rollout variation around the group mean~\cite{bardes2021vicreg}, while effective rank characterizes how this variation is distributed across principal directions~\cite{rankme}.

Building on this spectral view, we propose \textbf{SpeCal}, a two-level spectral calibration framework for unsupervised RLVR. Rather than constructing another pseudo-reward, SpeCal uses hidden-state geometry as complementary structural evidence for calibrating the advantages induced by an existing unlabeled reward scheme. 
\textbf{At the group level}, it combines spectral mass with an effective-rank-based band-pass gate to modulate the contribution of the entire rollout group according to the magnitude and directional organization of its representational variation. This allows SpeCal to distinguish informative rollout structures from collapsed or fragmented representation patterns, even when their endpoint statistics appear similar.
\textbf{At the rollout level}, SpeCal reuses the group effective rank to determine the number of dominant directions retained in a leave-one-out spectral subspace. For each rollout, the reference subspace is constructed from the remaining responses sampled for the same prompt, and the rollout is evaluated according to its conditional support within this cross-rollout structure. This design prevents a trajectory from being evaluated solely by the directions it creates itself, while measuring whether its representation is supported by the shared structure of the group.

We conduct extensive experiments on two 8B model backbones across six in-domain mathematical-reasoning benchmarks and three out-of-domain general-reasoning benchmarks. Averaged across the two backbones, SpeCal improves over the strongest unsupervised RLVR baselines by $2.7$ points in-domain and $1.1$ points out-of-domain, while matching or surpassing fully supervised RLVR in several settings. SpeCal consistently improves multiple unlabeled reward schemes, demonstrating that its gains are not tied to a particular pseudo-reward construction. Further ablations verify the complementary contributions of group-level spectral calibration and rollout-level conditional support. These results show that hidden-state spectral geometry provides an effective and general reliability signal for learning from self-generated supervision.

\section{Preliminary}

\subsection{Standard RLVR Paradigm}
Reinforcement learning with verifiable rewards (RLVR) optimizes language models using outcome rewards provided by rule-based verifiers.
Formally, given a labeled dataset $\mathcal{D}_{\mathrm{L}}$ of question--answer pairs $(q,a)$, the policy $\pi_\theta$ generates a response
$y \sim \pi_\theta(\cdot \mid q)$ for each question $q$.
Let $\hat{a}$ denote the final answer extracted from $y$,
the verifier assigns a binary reward
$R(y,a)=\mathbb{I}[\hat{a}=a]$,
where $\mathbb{I}[\cdot]$ is the indicator function.
We adopt the widely used \textit{Group Relative Policy Optimization}
(GRPO)~\citep{grpo} as our base algorithm. GRPO eliminates the need for a learned value model by estimating advantages from the rewards of multiple responses to the same question.
Specifically, for each question $q$, we sample $G$ responses
$\{y_j\}_{j=1}^{G}$ from the old policy
$\pi_{\theta_{\mathrm{old}}}$ and compute their rewards
$R(y_j,a)$.
The group-normalized advantage $\hat{A}_i$ is given by
\vspace{-0.1\baselineskip}
\begin{equation}
\hat{A}_j =
\frac{
R(y_j,a)
-
\text{mean}\!\left(
    \{R(y_j,a)\}_{j=1}^{G}
\right)
}{
\text{std}\!\left(
    \{R(y_j,a)\}_{j=1}^{G}
\right)
}.
\label{eq:grpo_advantage}
\end{equation}
Then, the GRPO objective is defined as
\begin{equation}
\scalebox{0.94}{$
\begin{aligned}
    \mathcal{J}_{\text{GRPO}}
    & (\theta; \mathcal{D})
    = \mathbb{E}[q\sim\mathcal{D}, \{y_j\}_{j=1}^G\sim\pi_{\theta_\text{old}}(\cdot \mid q)] \\
    &\hspace{-2em} \frac{1}{G} \! \sum_{j=1}^G \! \frac{1}{|y_j|} \! \sum_{t=1}^{|y_j|} \! \text{CLIP}\bigl(\gamma_{j,t}(\theta), \hat{A}_j, \epsilon\bigr) \! - \! \beta \! \cdot \! \mathbb{D}_{\text{KL}}[\pi_\theta \,\|\, \pi_{\text{ref}}].
\end{aligned}
$}
\label{eq:GRPO_loss}
\end{equation}
Here,
${\gamma_{j,t}(\theta) \!=\! {\pi_\theta(y_{j,t} | q, y_{j,<t})}/{\pi_{\theta_{\text{old}}}(y_{j,t} | q, y_{j,<t})}}$ is the importance weight, $\text{CLIP}(\gamma, A, \epsilon) = \min[\gamma \!\cdot\! A, \text{clip}(\gamma; 1\!-\!\epsilon, 1\!+\!\epsilon) \!\cdot\! A]$ is the clipped surrogate objective, and $\mathbb{D}_{\text{KL}}$ denotes the KL divergence.

\subsection{Unsupervised RLVR} 
Unlike standard RLVR, unsupervised RLVR operates on an unlabeled question set
$\mathcal{D}_{\mathrm{U}}=\{q\}$ and derives rewards from the model's own
rollouts. For each question $q$, the policy samples $G$ responses
$\mathcal{Y}(q)=\{y_j\}_{j=1}^{G}$ and extracts their final answers
$\{\hat{a}_j\}_{j=1}^{G}$. Given a pseudo-label $a^{*}$ inferred from these
responses, the unlabeled reward is commonly defined as
$R_{\mathrm{U}}(y_j)=\mathbb{I}[\hat{a}_j=a^{*}]$.
A representative approach is majority voting~\citep{ttrl}, which constructs
the pseudo-label as
$a^{*}=\text{MAJ}(\hat{a}_1,\ldots,\hat{a}_G)$.
The resulting rewards are normalized as in Eq.~\eqref{eq:grpo_advantage} and
optimized using the GRPO objective in Eq.~\eqref{eq:GRPO_loss}.
More generally, SpeCal is agnostic to how the pseudo-label or unlabeled reward
is constructed. Since these rewards are inferred from self-generated outcomes
rather than external verification, their induced advantages may have
heterogeneous reliability, which SpeCal calibrates without modifying the underlying reward.

\section{Methods}
\label{sec:method}
\begin{figure*}[!t]
  \centering
  \includegraphics[width=0.85\linewidth]{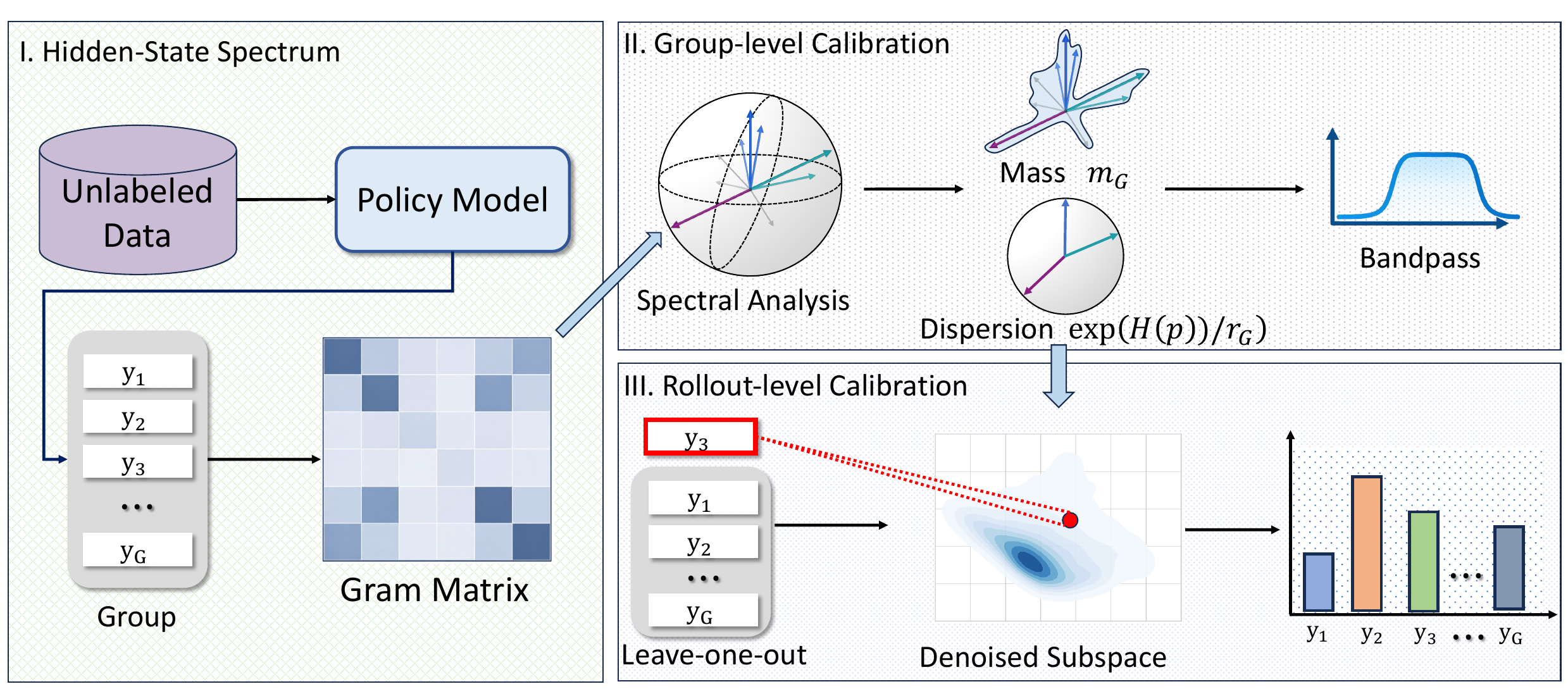}
   \caption{Overview of SpeCal. Given a rollout group, SpeCal constructs a centered Gram matrix from the hidden-state representations. Group-level calibration combines spectral mass $m_G$ with a band-pass gate over normalized effective rank $\eta_G$, while rollout-level calibration scores each rollout against the leave-one-out denoised subspace formed by the remaining rollouts. The resulting weights $w_G$ and $w_{R,j}$ calibrate the original unsupervised advantages without modifying the reward.}
   \Description{SpeCal calibrates rollout groups using hidden-state spectra and calibrates individual rollouts using leave-one-out spectral support.}
   \label{fig:method}
\end{figure*}

In this section, we introduce \text{SpeCal}, a fine-grained bi-level calibration framework for unsupervised RLVR. As shown in Fig.~\ref{fig:method}, \text{SpeCal} evaluates the balanced spectral geometry across rollouts at the group level, and measures each trajectory's conditional spectral support against the leave-one-out subspace at the rollout level.

\subsection{Group Calibration via Spectral Geometry }
\subsubsection{Response Representations and Group Spectrum.}
For the $j$-th rollout $y_j$ with $T_j$ generated tokens, let
$\bm{h}_{j,t}^{(\ell)}\in\mathbb{R}^{d}$ denote the hidden state of token $t$
at layer $\ell$. We aggregate information across a selected set of layers
$\mathcal{L}$ and then mean-pool over all generated tokens:
\begin{equation}
\tilde{\bm{h}}_j=\frac{1}{T_j}\sum_{t=1}^{T_j}\text{Concat}_{\ell\in\mathcal{L}}
\bm{h}_{j,t}^{(\ell)}.
\label{eq:cross_layer_response_repr}
\end{equation}
Mean pooling is motivated by the observation that semantic information is
distributed throughout the generated sequence, making the aggregated
representation more informative than relying on any single token
state~\citep{tokenmean}. We then obtain the normalized response representation
$\bm{z}_j=\tilde{\bm{h}}_j/\lVert\tilde{\bm{h}}_j\rVert_2$ and stack all $G$ representations into the group matrix
$\mathbf{Z}=[\bm{z}_1^\top;\ldots;\bm{z}_G^\top]
\in\mathbb{R}^{G\times|\mathcal{L}|d}$.
To isolate variation across rollouts, we center the group representations
$\mathbf{Z}_c$ and construct their Gram matrix $\mathbf{K}_G$:
\begin{equation}
\mathbf{Z}_{c}=\left(
\mathbf{I}_G-\frac{1}{G}\mathbf{1}_G\mathbf{1}_G^{\top}\right)\mathbf{Z},\qquad
\mathbf{K}_{G}=\mathbf{Z}_{c}\mathbf{Z}_{c}^{\top}\in\mathbb{R}^{G\times G}.
\label{eq:group_gram_matrix}
\end{equation}
Since $\mathbf{K}_{G}$ is positive semidefinite and $\mathbf{Z}_{c}$ is
centered, its rank is at most $G-1$. We denote the group spectral dimension as
$r_G=\min(G-1,|\mathcal{L}|d)$, and write the corresponding eigenvalues as
$\lambda_1\geq\cdots\geq\lambda_{r_G}\geq 0$, including zero-valued directions
when the effective rank is smaller than $r_G$. These spectral quantities are
used below to define the group-level calibration score.

\subsubsection{Group-Level Spectral Calibration.}
A rollout group provides a useful learning signal only when it offers nontrivial yet coherent semantic variation. If the responses collapse to a small region in representation space, the group lacks semantic contrast for relative comparison; if the responses disperse into too many weakly related modes, the variation may instead reflect outliers or fragmented reasoning paths. This motivates a calibration score that considers both the amount of inter-rollout variation and the effective dimensionality of this variation.

Given the group spectrum $\{\lambda_k\}_{k=1}^{r_G}$, we first compute the spectral mass $m_G=\frac{1}{G}\sum_{k=1}^{r_G}\lambda_k$, which measures the overall dispersion of the rollout representations. We then normalize the eigenspectrum as $p_k=\frac{\lambda_k}{\sum_{\ell=1}^{r_G}\lambda_{\ell}}$ and compute the entropy effective rank~\cite{roy2007effective}:
\begin{equation}
\operatorname{ER}_G=\exp(-\sum_{k=1}^{r_G}p_k\log p_k).
\label{eq:group_effective_rank}
\end{equation}
The normalized effective rank $\eta_G=\operatorname{ER}_G/r_G$ estimates the fraction of available spectral directions that are effectively used by the group~\cite{rankme}.

We do not assume that a larger effective rank is always preferable. Very small $\eta_G$ indicates collapse or single-axis variation, whereas overly large $\eta_G$ may correspond to fragmented groups with weak shared structure. We therefore use a Gaussian band-pass gate over $\eta_G$ and modulate the spectral mass by this gate:
\begin{equation}
w_G(q)=m_G\cdot \exp\!\left(-\frac{(\eta_G-\mu_G)^2}{2\tau_G^2}\right).
\label{eq:group_er_gated_mass}
\end{equation}
Here, $\mu_G$ denotes the target normalized effective rank and $\tau_G$ controls the tolerance of effect rank. Thus, $w_G(q)$ becomes large only when the rollout group has sufficient spectral mass and its effective dimensionality lies in the desired range. 
The unnormalized effective rank $\operatorname{ER}_G$ also provides the spectral bandwidth used in trajectory-level calibration. We set $k_G=\lceil \operatorname{ER}_G\rceil$, which estimates how many dominant semantic directions should be retained when constructing leave-one-out subspaces below.

\subsection{Rollout Calibration via Spectral Support}

\subsubsection{Denoised Leave-One-Out Subspace.}
The group-level weight $w_G(q)$ calibrates the overall usefulness of a rollout group, but it does not specify how the learning signal should be distributed among individual rollouts. For rollout-level calibration, a response should not certify itself by the spectral directions it alone creates: unusual hidden variation may arise from noise, verbosity, or idiosyncratic reasoning rather than reliable semantic structure~\citep{kool2019buy,ahmadian2024back}. We therefore evaluate each trajectory against a reference geometry independently formed by the remaining rollouts. Specifically, for the $j$-th trajectory, we remove $\bm{z}_j$ from the group and write $\mathbf{Z}_{-j}=[\bm{z}_1^\top;\ldots;\bm{z}_{j-1}^\top;\bm{z}_{j+1}^\top;\ldots;\bm{z}_G^\top]$. Let $\bar{\bm{z}}_{-j}=\frac{1}{G-1}\sum_{k\neq j}\bm{z}_k$ be the leave-one-out mean, and define the centered leave-one-out matrix as $\mathbf{Z}_{-j,c}=\mathbf{Z}_{-j}-\mathbf{1}\bar{\bm{z}}_{-j}^{\top}$.

To denoise this reference geometry, we do not use the full span of the remaining rollouts, since weak spectral directions may correspond to accidental or unstable variation. Instead, we retain only the dominant spectral directions that capture the effective semantic structure of the rollout group. Specifically, we compute the compact SVD of the centered leave-one-out matrix:
\begin{equation}
\mathbf{Z}_{-j,c}=\mathbf{U}_{-j}\operatorname{diag}(\sqrt{\mu_1},\ldots,\sqrt{\mu_r})\mathbf{V}_{-j}^{\top}.
\label{eq:loo_svd}
\end{equation}
Here, $\mu_1\geq\cdots\geq\mu_r>0$, and the columns of $\mathbf{V}_{-j}$ represent the spectral directions supported by the remaining rollouts. 
The group-level effective rank provides an adaptive spectral bandwidth for this leave-one-out subspace: we retain the top $k_G=\lceil\operatorname{ER}_G\rceil$ available directions from $\mathbf{V}_{-j}$ and denote the truncated basis by $\mathbf{V}_{-j}^{(k_G)}$. The resulting denoised leave-one-out projector is

\begin{equation}
\mathbf{P}_{-j}^{(k_G)}=\mathbf{V}_{-j}^{(k_G)}\mathbf{V}_{-j}^{(k_G)\top}.
\label{eq:loo_topk_projector}
\end{equation}
This construction preserves the dominant directions supported by the remaining rollouts while discarding weaker directions beyond the group-level effective dimensionality. In this way, leave-one-out prevents self-certification, and effective-rank truncation prevents the trajectory from being evaluated against noisy or accidental spectral directions.

\subsubsection{Conditional Spectral Support.}
Given the effective-rank truncated leave-one-out projector $\mathbf{P}_{-j}^{(k_G)}$, we measure how much of the held-out rollout is supported by the semantic structure established by the remaining rollouts. Specifically, using the centered rollout representation $\bm{x}_j=\bm{z}_j-\bar{\bm{z}}_{-j}$, we define the conditional spectral support score as:
\begin{equation}
s_{R,j}=\frac{\bm{x}_j^{\top}\mathbf{P}_{-j}^{(k_G)}\bm{x}_j}{\lVert \bm{x}_j\rVert_2^2}.
\label{eq:conditional_spectral_support}
\end{equation}
The numerator measures the projected energy of $\bm{x}_j$ onto the effective leave-one-out spectral subspace~\cite{ndiour2022subspace}, while the denominator measures its total energy relative to the same reference mean. Therefore, $s_{R,j}$ measures conditional support rather than intrinsic trajectory complexity: a trajectory receives a high score only when its group-relative variation is explained by dominant directions independently supported by the remaining rollouts.

We then normalize the support scores within each rollout group and apply clipping to obtain the trajectory-level calibration weight:
\begin{equation}
w_{R,j}=\operatorname{clip}\!\left(\frac{s_{R,j}}{\frac{1}{G}\sum_{k=1}^{G}s_{R,k}},w_R^{-},w_R^{+}\right).
\label{eq:trajectory_weight}
\end{equation}
Here, $w_R^{-}$ and $w_R^{+}$ are clipping bounds. The group-wise normalization keeps the average trajectory weight close to one before clipping, so $w_{R,j}$ redistributes learning emphasis among rollouts within the same group rather than changing the overall strength of the prompt-level update.

\subsubsection{Bi-Level Calibrated Objective.}
We finally combine the two calibration levels by reweighting the original unsupervised advantage. Let $w_G(q)$ denote the group-level calibration coefficient and $w_{R,j}$ denote the trajectory-level calibration coefficient. The calibrated advantage is defined as
\begin{equation}
A^{\mathrm{cal}}_{q,j}=\mathcal{N}_b(\operatorname{sg}(w_G(q)))\cdot\mathcal{N}_g\!\left(\operatorname{sg}(w_{R,j})\hat{A}^{\mathrm{U}}_{q,j}\right).
\label{eq:SpeCal_advantage}
\end{equation}
Here, $\operatorname{sg}(\cdot)$ denotes stop-gradient, and $\mathcal{N}_b(\cdot),\mathcal{N}_g(\cdot)$ denotes RMS normalization~\cite{rms} within the batch and rollout group respectively, which preserves bi-level advantages scale after reweighting. Thus, $w_G(q)$ controls how strongly the whole rollout group contributes to learning, while $w_{R,j}$ only redistributes learning emphasis among trajectories within the group.
The policy is then optimized with the same GRPO objective as Eq.~\eqref{eq:GRPO_loss}, replacing $\hat{A}_j$ with $A^{\mathrm{cal}}_{q,j}$:
\begin{equation}
\mathcal{J}_{\text{SpeCal}}(\theta;\mathcal{D}_{\mathrm{U}})
=\mathcal{J}_{\mathrm{GRPO}}(\theta;\mathcal{D}_{\mathrm{U}})\big|_{\hat{A}_{q,j} \leftarrow A^{\mathrm{cal}}_{q,j}}.
\label{eq:SpeCal_objective}
\end{equation}


\section{Theoretical Analysis}
\label{sec:theory_overview}

We analyze SpeCal from two complementary perspectives. First, we characterize the information limitation of endpoint-based unsupervised signals. Second, we show that the spectral quantities used by SpeCal estimate a well-defined marginal geometry of the rollout process. Formal assumptions and proofs are deferred to Appendix~\ref{app:theory}.

\subsection{Information Loss of Endpoint Signals}
Let $\mathcal{X}_q=\{(\hat{a}_j,\bm{z}_j)\}_{j=1}^{G}$ denote the complete rollout evidence for question $q$, including both decoded outcomes $\hat{a}$ and response representations $\bm{z}$. Existing unsupervised methods typically construct their learning signals from an endpoint summary $E_q=\Phi(\hat{a}_1,\ldots,\hat{a}_G)$. The following proposition formalizes the information loss induced by this compression.

\begin{proposition}[Endpoint Information Loss]
\label{prop:endpoint_information_loss}
For any random variable $C_q$ describing the reliability of the induced supervision signal,
\begin{equation}
I(C_q;\mathcal{X}_q)\geq I(C_q;E_q).
\label{eq:endpoint_information_loss}
\end{equation}
The gap is strict whenever endpoint-equivalent rollout groups can have different conditional supervision reliability.
\end{proposition}

Proposition~\ref{prop:endpoint_information_loss} shows that the limitation of endpoint-based methods is not restricted to a particular reward design: once the rollout group is compressed into decoded statistics, reliability-relevant representational information may become irrecoverable.

\subsection{Marginal Spectral Consistency}

For a fixed question $q$ and policy $\pi$, let $\bm{z}$ and $\bm{z}'$ be two i.i.d. response representations sampled from $P_{\pi}(\bm{z}\mid q)$. We define the marginal inter-rollout geometry as $\mathbf{\Sigma}^{\mathrm{marg}}_q=\frac{1}{2}\mathbb{E}[(\bm{z}-\bm{z}')(\bm{z}-\bm{z}')^\top\mid q]$, and estimate it using the centered empirical matrix $\widehat{\mathbf{\Sigma}}_{q,G}=\frac{1}{G-1}\mathbf{Z}_c^\top\mathbf{Z}_c$. To distinguish finite-group estimates from their marginal counterparts, we use $\widehat{w}_G(q)$ and $\widehat{s}_{R,j}$ to denote the group weight and conditional support computed by SpeCal from the sampled rollout group; these correspond to $w_G(q)$ and $s_{R,j}$ in Sec.~\ref{sec:method}. We denote by $w_G^\star(q)$ and $s_R^\star(\bm{z}_j)$ the same spectral constructions evaluated on the marginal geometry.

\begin{proposition}[Marginal Spectral Consistency]
\label{prop:marginal_spectral_consistency}
Under the regularity and rank-stability conditions stated in Appendix~\ref{app:theory},
\begin{equation}
\mathbb{E}[\widehat{\mathbf{\Sigma}}_{q,G}\mid q]=\mathbf{\Sigma}^{\mathrm{marg}}_q,
\bigl(\widehat{w}_G(q),\widehat{s}_{R,j}\bigr)
\xrightarrow[G\rightarrow\infty]{\mathrm{a.s.}}
\bigl(w_G^\star(q),s_R^\star(\bm{z}_j)\bigr).
\label{eq:marginal_spectral_consistency}
\end{equation}
\end{proposition}

Proposition~\ref{prop:marginal_spectral_consistency} shows that the finite-group calibration signals used by SpeCal converge to well-defined properties of the marginal rollout geometry, rather than depending on arbitrary spectral fluctuations of a particular sampled group.





\section{Experiments}
\subsection{Setup}
\subsubsection{Implementation Details.}
Our implementation is built upon the \text{verl} framework~\cite{verl} and follows the standard GRPO training recipe~\cite{grpo}. We conduct training on the challenging subset of DeepMath-103K~\cite{deepseekmath} with difficulty scores $\geq 6$. To evaluate the generality of SpeCal across different model families and initial reasoning capabilities, we experiment with backbones from both the Qwen and Llama families, including Qwen3-8B-Base and DeepSeek-R1-Distill-Llama-8B. All experiments are performed on 8 NVIDIA A100 GPUs with a total batch size of 128, a micro-batch size of 32, and a learning rate of $1\times10^{-6}$. For each prompt, we sample $G=8$ rollouts during training. To construct hidden-state representations, we uniformly select eight layers across the depth of the policy model. For group-level calibration, we set the target normalized effective rank to $\mu_G=0.45$ and its tolerance to $\tau_G=0.25$. For rollout-level calibration, the conditional support weights are clipped to $[0.25,2.0]$. We use the same SpeCal hyperparameters across all backbones and settings. Additionally, unless otherwise specified, all analyses in this section are conducted
on Qwen3-8B-Base using the default training and evaluation configurations. Additional implementation details are provided in
Appendix~\ref{app:exp_setup}.

\subsubsection{Evaluation.}
Following the evaluation protocols adopted in prior work~\cite{yan2026learning,yang2025trapo,zhu2026geomin}, we conduct comprehensive evaluations on a diverse collection of mathematical and general reasoning benchmarks. For mathematical reasoning, we consider six competition-level datasets: AIME 2024, AIME 2025, MATH-500~\cite{hendrycks2021measuring}, Minerva~\cite{lewkowycz2022solving}, AMC~\cite{li2024numinamath}, and OlympiadBench~\cite{he2024olympiadbench}. To evaluate out-of-domain generalization, we further include three general reasoning benchmarks: ARC-c~\cite{clark2018think}, GPQA-Diamond~\cite{rein2023gpqa} (denoted as GPQA), and MMLU-Pro~\cite{wang2024mmlu}. To account for differences in test-set size, we report avg@32 for AIME 2024/2025 and AMC, pass@1 for MMLU-Pro, and avg@4 for the remaining benchmarks. All evaluations use a temperature of 0.6 and top-$p$ of 1.0 under a consistent decoding configuration.

\subsubsection{Baselines.}We compare SpeCal with six representative unsupervised RLVR baselines spanning consensus-based pseudo-labeling, entropy and confidence based intrinsic rewards, and teacher-guided supervision. (1) \textbf{TTRL}~\cite{ttrl} takes the majority-voted answer within each rollout group as the pseudo-label and assigns positive rewards to responses that match it. (2) \textbf{Tok-entropy}~\cite{entropy_base} uses the average token entropy as an intrinsic confidence signal, favoring rollouts with more concentrated token-level predictions. (3) \textbf{Seq-entropy}~\cite{entropy_base} measures confidence at the sequence level and rewards responses with higher overall generation likelihood. (4) \textbf{Self-certainty}~\cite{zhao2025learning} quantifies intrinsic certainty through the KL divergence between the predicted token distributions and the uniform distribution, assigning larger rewards to more decisive rollouts. (5) \textbf{SCRL}~\cite{scrl} improves the robustness of consensus-based supervision through selective-complementary reward construction: it applies strict consensus criteria to retain reliable positive pseudo-labels and introduces entropy-gated negative pseudo-labeling to suppress likely incorrect trajectories when the answer distribution is dispersed. (6) \textbf{Co-rewarding}~\cite{coreward} maintains a slowly updated reference teacher and constructs rewards according to the agreement between sampled responses and teacher-generated pseudo-labels. For a fair comparison, all baselines are implemented using the same model backbones, training data, rollout budget, and GRPO optimization configuration. About other related work mentioned above like RESTRAIN~\citep{yu2026restrain} and DARE~\citep{dare}, we did not make a comparison because the code has not yet been open-sourced.

\subsection{Main Results}

\begin{table*}[!t]
\centering
\caption{Main results on Qwen3-8B-Base and
DeepSeek-R1-Distill-Llama-8B.
Each entry reports the mean accuracy over three independent runs with different random seeds.
ID Avg. is computed over six mathematical-reasoning benchmarks,
while OOD Avg. is computed over three general-reasoning benchmarks.
\textbf{Bold} marks the highest mean among unsupervised methods,
while gray rows provide the before-RL and fully supervised references.}
\label{tab:main_results}
\setlength{\tabcolsep}{3pt}
\renewcommand{\arraystretch}{1.2}
\resizebox{\textwidth}{!}{%
\begin{tabular}{lcccccc|cccc}
\toprule
\multirow{2}{*}{\textbf{Methods}} & \multicolumn{6}{c}{\textbf{In-Domain Performance}} & \multicolumn{4}{c}{\textbf{Out-of-Domain Performance}} \\
\cmidrule(lr){2-7} \cmidrule(lr){8-11}
 & \textbf{AIME 24/25} & \textbf{AMC} & \textbf{MATH-500} & \textbf{Minerva} & \textbf{Olympiad} & \textbf{Avg.} & \textbf{ARC-c} & \textbf{GPQA} & \textbf{MMLU-Pro} & \textbf{Avg.} \\
\midrule
\multicolumn{11}{c}{\textbf{Qwen3-8B-Base}} \\
\midrule
Before RL & 9.4/7.3 & 38.0 & 64.4 & 25.6 & 32.5 & 29.5 & 29.7 & 11.5 & 46.6 & 29.3 \\
\textcolor{gray}{Fully Supervised} & \textcolor{gray}{23.1/19.8} & \textcolor{gray}{55.1} & \textcolor{gray}{85.0} & \textcolor{gray}{46.5} & \textcolor{gray}{50.9} & \textcolor{gray}{46.7} & \textcolor{gray}{93.9} & \textcolor{gray}{50.5} & \textcolor{gray}{65.5} & \textcolor{gray}{70.0} \\
\midrule
Tok-entropy & 12.3/11.1 & 44.8 & 78.5 & 34.6 & 41.8 & 37.2 & 43.1 & 10.2 & 56.7 & 36.7 \\
Seq-entropy & 14.2/15.8 & 51.9 & 78.3 & 38.0 & 43.1 & 40.2 & 81.3 & 20.8 & 59.5 & 53.9 \\
TTRL & 17.7/12.5 & 54.4 & 81.7 & 43.8 & 46.0 & 42.7 & 91.3 & 43.6 & 63.0 & 66.0 \\
Self-certainty & 9.1/7.7 & 45.2 & 77.1 & 41.1 & 39.7 & 36.6 & 64.4 & 18.9 & 59.2 & 47.5 \\
SCRL & 16.7/14.2 & 49.4 & 76.2 & 40.4 & 40.2 & 39.5 & 90.2 & 41.6& 62.1 & 64.6 \\
Co-rewarding & 14.4/12.2 & 52.3 & 79.2 & 43.1 & 44.4 & 40.9 & 91.0 & 27.8 & 62.7 & 60.5 \\
\rowcolor{Gray} \textbf{SpeCal (ours)} & \textbf{20.9}/\textbf{16.3} & \textbf{58.1} & \textbf{83.9}& \textbf{44.7} & \textbf{49.5} & \textbf{45.6} & \textbf{91.7} & \textbf{46.7} & \textbf{63.4} & \textbf{67.3} \\
\midrule
\multicolumn{11}{c}{\textbf{DeepSeek-R1-Distill-Llama-8B}} \\
\midrule
Before RL &13.3/23.3  &50.6  &72.6  &26.1  &40.4  &37.7  &5.3  &9.3  &42.1  &18.9  \\
\textcolor{gray}{Fully Supervised} & \textcolor{gray}{43.3/23.3}  & \textcolor{gray}{66.3} & \textcolor{gray}{82.2} & \textcolor{gray}{36.8} & \textcolor{gray}{54.7} & \textcolor{gray}{51.1} & \textcolor{gray}{46.9} & \textcolor{gray}{37.3} & \textcolor{gray}{51.2} & \textcolor{gray}{45.1} \\
\midrule
Tok-entropy & 3.3/3.3 & 36.1 & 48.6 & 15.4 & 27.6 & 22.4 & 34.3 & 7.6 & 25.9 & 22.6 \\
Seq-entropy & 6.7/3.3 & 28.9 & 48.6 & 15.8 & 29.4 & 22.1 & 40.3 & 14.3 & 27.4 & 27.3 \\
TTRL &\textbf{33.3}/16.7 & 66.3 & 82.2 & 34.6 & 54.5 & 47.9 & 23.8 & 36.9 & 48.7 &36.5  \\
Self-certainty & 16.7/16.7 & 51.8 & 73.0 & 26.1 & 43.8 & 38.0 & 25.3 & 16.2 & 44.1 & 28.5 \\
SCRL & 21.7/20.5 & 61.4 & 77.8 & 31.3 & 43.1 & 42.6 & \textbf{58.7} & 40.1 & 51.2 & 50.0 \\
Co-rewarding & 30.0/20.0 & \textbf{69.9} & 79.6 & 28.7 & 48.1 & 46.1 & 31.3 & 29.9 & 51.2 & 37.5 \\
\rowcolor{Gray} \textbf{SpeCal (ours)} & \textbf{33.3}/\textbf{23.3} & \textbf{69.9} & \textbf{83.6}& \textbf{35.5} & \textbf{55.9} & \textbf{50.3} & 58.1 & \textbf{40.7} & \textbf{53.9} & \textbf{50.9} \\
\bottomrule
\end{tabular}%
}
\end{table*}
Table~\ref{tab:main_results} compares SpeCal with six representative unsupervised RLVR baselines on two 8B model backbones. We report mathematical reasoning tasks and out-of-domain generalization tasks separately, while including the before-RL and fully supervised results as reference points.

SpeCal delivers consistent in-domain improvements across both model families. On Qwen3-8B-Base, it raises the average accuracy from $42.7$ with TTRL to $45.6$, corresponding to a $2.9$-point absolute gain and a $6.8\%$ relative improvement. The gain is broad across tasks: SpeCal improves all six mathematical-reasoning benchmarks, with increases ranging from $0.9$ points on Minerva to $3.8$ points on AIME 2025. A similar pattern is observed on DeepSeek-R1-Distill-Llama-8B, where SpeCal improves the strongest unsupervised average from $47.9$ to $50.3$. It improves five benchmarks over TTRL and ties it on AIME 2024. Averaged across the two backbones, SpeCal improves the strongest unsupervised baseline by $2.7$ points in-domain. In comparison, the entropy-based methods vary considerably across model families and reduce the DeepSeek model below its before-RL performance, highlighting the difficulty of transferring a fixed intrinsic reward heuristic across different policy models.

SpeCal also achieves the best unsupervised OOD average on both model backbones. On Qwen3-8B-Base, it reaches an OOD average of $67.3$, outperforming TTRL at $66.0$ by $1.3$ points. SpeCal improves over TTRL on all three OOD benchmarks, with gains of $0.4$ points on ARC-c, $3.1$ points on GPQA, and $0.4$ points on MMLU-Pro. On DeepSeek-R1-Distill-Llama-8B, SpeCal obtains an OOD average of $50.9$, exceeding the strongest baseline, SCRL at $50.0$, by $0.9$ points. At the task level, it outperforms the strongest baseline by $0.6$ points on GPQA and $2.7$ points on MMLU-Pro, while trailing SCRL by $0.6$ points on ARC-c. Averaged across the two backbones, SpeCal improves the strongest unsupervised OOD baseline by $1.1$ points. These results demonstrate consistent average-level gains across model families, while also showing that the task-level advantage depends on the target distribution.

Notably, SpeCal approaches or even surpasses fully supervised RLVR in several settings. Its in-domain average is only $1.1$ points below the fully supervised reference on Qwen3-8B-Base and $0.8$ points below it on DeepSeek-R1-Distill-Llama-8B. SpeCal outperforms the supervised reference on AMC with Qwen3-8B-Base. On DeepSeek-R1-Distill-Llama-8B, it exceeds the supervised reference on MATH-500 and OlympiadBench and matches it on AIME 2025 and AMC. SpeCal also surpasses fully supervised RLVR on all three OOD benchmarks with the Llama backbone. These results highlight the strong competitiveness of SpeCal and the potential of unsupervised RLVR as a scalable alternative when manually annotated rewards are unavailable.

\subsection{Further Analysis}
\subsubsection{Ablation Study.}

\begin{table*}[!t]
\centering
\caption{
Ablation study of SpeCal components.
$m_G$ denotes the group-level spectral mass component, $\eta_G$ denotes the normalized effective rank used by the group-level Gaussian gate, and $w_R$ denotes trajectory-level conditional spectral support.
}
\label{tab:ablation_components}
\setlength{\tabcolsep}{4pt}
\renewcommand{\arraystretch}{1.15}
\resizebox{\textwidth}{!}{%
\begin{tabular}{lccc|ccccccc}
\toprule
\multirow{2}{*}{\textbf{Method}} 
& \multicolumn{3}{c|}{\textbf{Calibration Components}} 
& \multicolumn{7}{c}{\textbf{Benchmarks}} \\
\cmidrule(lr){2-4} \cmidrule(lr){5-11}
& \abcell{$m_G$} & \abcell{$\eta_G$} & \abcell{$w_R$}
& \textbf{AIME24} & \textbf{AIME25} & \textbf{AMC} & \textbf{MATH-500} & \textbf{Minerva} & \textbf{Olympiad} & \textbf{Avg.} \\
\midrule
Majority Vote RLVR & \abcell{\xmark} & \abcell{\xmark} & \abcell{\xmark} & \textcolor{gray}{17.7} & \textcolor{gray}{12.5} & \textcolor{gray}{54.4} & \textcolor{gray}{81.7} & \textcolor{gray}{43.8} & \textcolor{gray}{46.0} & \textcolor{gray}{42.7}
\\
SpeCal w/o Spread & \abcell{\xmark} & \abcell{\cmark} & \abcell{\cmark} & 19.2 & 15.7 & 56.0 & 82.0 & 44.3 & \textbf{49.7} & 44.5 \\
SpeCal w/o ER Gate & \abcell{\cmark} & \abcell{\xmark} & \abcell{\cmark} & 19.2  & 13.8 & 57.6 & 81.6 & 44.5 & 48.7 & 44.2 \\
SpeCal w/o Traj. Support & \abcell{\cmark} & \abcell{\cmark} & \abcell{\xmark} & 20.4 & 13.6 & 53.9 & 82.5 & \textbf{46.9} & 48.8 & 44.3 \\
\rowcolor{Gray}\textbf{SpeCal} & \abcell{\cmark} & \abcell{\cmark} & \abcell{\cmark} & \textbf{20.9} & \textbf{16.3} & \textbf{58.1} & \textbf{83.9}& 44.7 & 49.5 & \textbf{45.6} \\
\bottomrule
\end{tabular}%
}
\end{table*}

We ablate SpeCal according to its natural bi-level decomposition: group-level calibration contains two spectral geometry factors, while trajectory-level calibration controls within-group reweighting. At the group level, $m_G$ measures the spectral mass, i.e., how much the rollout representations spread from their center, and $\eta_G$ measures the effective-rank structure, i.e., how this variation is distributed across spectral directions. Accordingly, \text{SpeCal w/o Spread} removes the spectral mass component $m_G$, and \text{SpeCal w/o ER Gate} removes the effective-rank gate over $\eta_G$. At the trajectory level, \text{SpeCal w/o Traj. Support} removes $w_R$, disabling conditional spectral support while retaining group-level calibration. \text{Majority Vote RLVR} removes all spectral calibration, and \text{SpeCal} enables all three components.

As shown in Table~\ref{tab:ablation_components}, every partial SpeCal variant improves the majority-vote baseline, raising the average score from $42.7$ to $44.2$--$44.5$. However, removing $m_G$, $\eta_G$, or $w_R$ decreases the full SpeCal performance by $1.1$, $1.4$, and $1.3$ points, respectively, indicating that the three components provide complementary information. Notably, removing spectral mass and removing effective rank produce similar averages, but lead to different task-wise behaviors: the former performs better on AIME25 and Olympiad, whereas the latter is stronger on AMC and Minerva. This contrast suggests that the magnitude and directional organization of group variation capture distinct properties rather than acting as interchangeable measures.
Trajectory-level support yields the largest average contribution. Compared with the group-only variant, adding $w_R$ improves AIME25 and AMC by $2.7$ and $4.2$ points, respectively, showing the benefit of distinguishing trajectories that are supported by the shared spectral structure from those dominated by idiosyncratic directions. Although individual ablations obtain slightly better results on Minerva or Olympiad, none is consistently superior across tasks: the complete SpeCal achieves the best average performance and the best result on four of the six benchmarks. Overall, the results show that spectral mass determines whether sufficient group-level variation exists, effective rank characterizes how this variation is organized, and conditional support further allocates learning emphasis within the group.

\subsubsection{Compatibility with Different Unlabeled Reward Schemes.}
\begin{figure}[t]
    \centering
    \subfloat[Comparison of ID Average\label{fig:reward_compat_id}]{%
        \includegraphics[width=0.8\columnwidth]
        {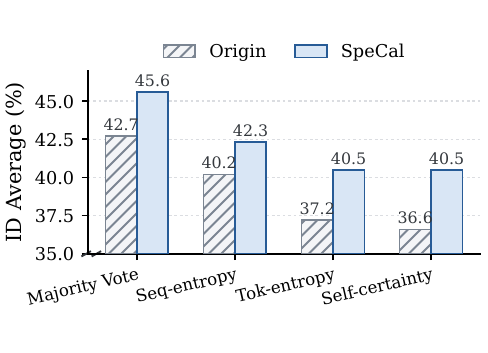}%
    }\\[-0.7em]
    \subfloat[Comparison of OOD Average\label{fig:reward_compat_ood}]{%
        \includegraphics[width=0.8\columnwidth]
        {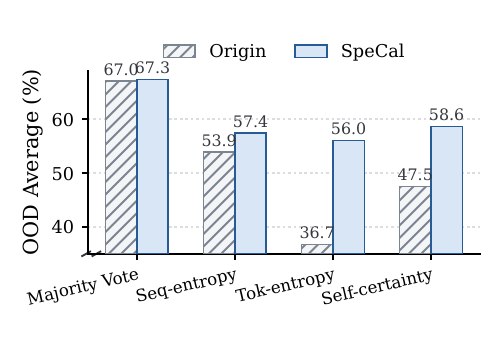}%
    }

    \caption{
    Compatibility of SpeCal with different unlabeled reward schemes.
    Gray bars denote the original schemes, while blue bars
    denote their SpeCal-enhanced variants.
    }
    \label{fig:reward_scheme_compatibility}
\end{figure}

We investigate whether SpeCal generalizes across different constructions of unlabeled rewards by applying it to Majority Vote, sequence-level entropy (Seq-entropy), token-level entropy (Tok-entropy), and Self-certainty. As shown in Fig.~\ref{fig:reward_scheme_compatibility}, SpeCal improves the ID average under all four schemes, yielding gains of $2.1$--$3.9$ percentage points. The OOD results reveal clearer differences: Majority Vote improves from $67.0$ to $67.3$, Seq-entropy from $53.9$ to $57.4$, Self-certainty from $47.5$ to $58.6$, and Tok-entropy from $36.7$ to $56.0$. A clear trend emerges from these results: reward schemes with lower original OOD performance obtain larger gains from SpeCal. Because each comparison keeps the underlying reward unchanged and differs only in whether its induced advantages are spectrally calibrated, the improvements demonstrate that SpeCal provides information complementary to the original reward signal. The modest gain on Majority Vote about OOD suggests that there is less room for improvement when the reward already transfers well, whereas the substantially larger gains on Tok-entropy and Self-certainty show that spectral calibration is especially useful when the original supervision transfers poorly. Overall, these results demonstrate that SpeCal is compatible with diverse unlabeled reward schemes and improves their robustness under distribution shift without redesigning the rewards themselves.

\subsubsection{Sensitivity Analysis.}

\begin{figure*}[ht]
    \centering
    \subfloat[Analysis on $\mu_G$\label{fig:sensi_1}]{%
        \includegraphics[width=0.24\textwidth]
        {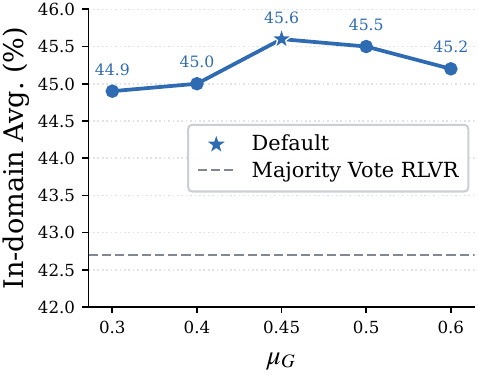}%
    }\hfill
    \subfloat[Analysis on $\tau_G$\label{fig:sensi_2}]{%
        \includegraphics[width=0.24\textwidth]
        {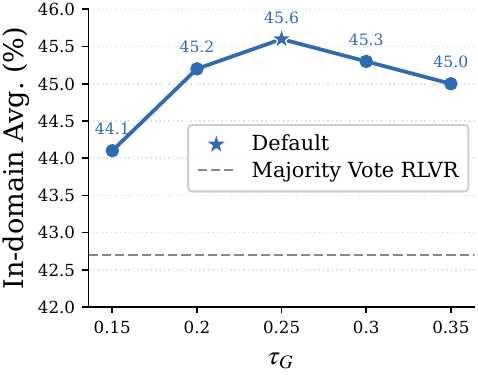}%
    }\hfill
    \subfloat[Analysis on $w_R^{-}$\label{fig:sensi_3}]{%
        \includegraphics[width=0.24\textwidth]
        {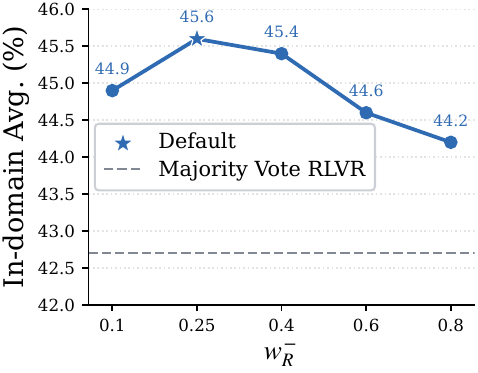}%
    }\hfill
    \subfloat[Analysis on $w_R^{+}$\label{fig:sensi_4}]{%
        \includegraphics[width=0.24\textwidth]
        {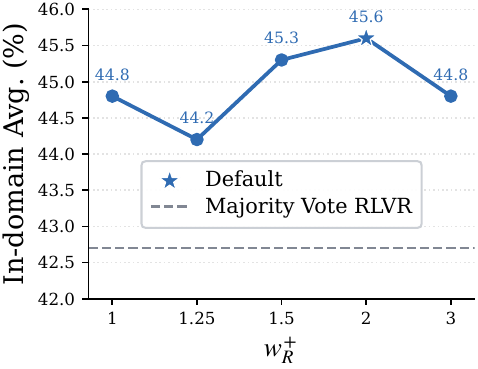}%
    }
    \caption{Hyperparameter sensitivity of SpeCal to the target normalized effective rank $\mu_G$, the effective-rank tolerance $\tau_G$, and the lower and upper
trajectory-weight clipping bounds $w_R^{-}$ and $w_R^{+}$.
The blue star marks the default setting in each sweep, while the dashed
horizontal line denotes the Majority Vote RLVR baseline.
}
    \Description{Four line plots report performance sensitivity to the group effective-rank center and tolerance and to the lower and upper rollout-weight clipping bounds.}
    \label{fig:hyper_sensitivity}
\end{figure*}
Figure~\ref{fig:hyper_sensitivity} evaluates the sensitivity of SpeCal to the target normalized effective rank $\mu_G$, its tolerance $\tau_G$, and the trajectory-weight clipping bounds $w_R^{-}$ and $w_R^{+}$. Across all four one-dimensional sweeps, the in-domain average ranges from $44.1$ to $45.6$, remaining at least $1.4$ points above the Majority Vote RLVR baseline of $42.7$. The default values $\mu_G=0.45$, $\tau_G=0.25$, $w_R^{-}=0.25$, and $w_R^{+}=2.0$ achieve the best observed score in their respective sweeps, while several neighboring settings remain competitive. 
At the group level, performance varies only moderately across $\mu_G\in[0.40,0.60]$, suggesting that useful rollout groups occupy a relatively broad effective-dimensionality regime rather than a sharply localized target. The tolerance $\tau_G$ has a more visible effect: an overly narrow tolerance assigns substantial weights to only a small range of effective-rank structures, reducing the effective utilization of otherwise informative rollout groups. Conversely, increasing $\tau_G$ beyond the default gradually weakens the selectivity of the Gaussian gate and makes spectrally distinct groups receive more similar weights.
At the trajectory level, the clipping bounds determine the dynamic range available for within-group calibration. Setting $w_R^{-}$ too low allows weakly supported trajectories to be strongly suppressed, whereas raising it toward $1$ compresses the weight distribution and limits the ability of SpeCal to distinguish trajectories with different support levels. The upper bound exhibits a complementary trade-off: a restrictive value leaves little room to emphasize highly supported trajectories, while an overly large value may concentrate the update on a small subset of rollouts. Although the upper-bound sweep is mildly non-monotonic, the best performance is obtained at the moderate setting $w_R^{+}=2.0$, with values around $1.5$--$2.0$ generally outperforming more restrictive or permissive choices. Overall, SpeCal remains effective across a broad range of configurations, while the selected defaults balance group-level selectivity, sample utilization, and sufficient trajectory-level weight contrast.

\section{Related Work}
\paragraph{\textbf{Unsupervised RLVR}}
Unsupervised RLVR derives supervision from model-generated responses without external labels. TTRL~\citep{ttrl} constructs pseudo-labels through majority voting, whereas entropy-based methods~\citep{entropy_base} and Intuitor~\citep{zhao2025learning} use intrinsic confidence signals. Co-rewarding stabilizes self-generated supervision through cross-question agreement or a slowly updated teacher~\citep{coreward}. More recent approaches retain richer endpoint information: distribution-aware reward estimation models answer-distribution statistics~\citep{dare}, while RESTRAIN combines vote-frequency weighting, negative advantage offsets, and question-level self-consistency~\citep{yu2026restrain}. Nevertheless, because these methods optimize rewards derived from the model's own outputs, errors in the proxy signals can be repeatedly reinforced during training. Agreement-based signals may stabilize an incorrect answer as the group consensus, producing spurious self-consistency, while confidence-based objectives may continue sharpening a distribution whose confidence is misaligned with correctness. Under prolonged optimization, the policy can exploit these proxy signals without improving task correctness, resulting in reward hacking and eventual training collapse~\citep{shafayat2025can,he2026far}. These findings highlight the fragility of relying exclusively on self-generated endpoint signals, whose reliability estimates remain derived from decoded frequencies, confidence, or teacher agreement rather than rollout representation geometry. \textsc{SpeCal} is complementary to these reward-construction strategies: it leaves the underlying reward unchanged and calibrates the induced advantages through group spectral geometry and rollout-level cross-rollout support.
\paragraph{\textbf{Spectral Methods}}
Spectral methods reveal global structure in high-dimensional representations. Spectral Contrastive Learning connects self-supervised objectives to augmentation-graph eigendecomposition~\citep{spectralcl}, singular-spectrum analyses diagnose dimensional collapse~\citep{dimcollapse}, and RankMe uses effective rank as a label-free representation-quality measure~\citep{rankme}. For LLM generations, INSIDE detects hallucinations using the covariance spectrum of hidden-state response embeddings~\citep{inside}, Spectral Uncertainty decomposes predictive uncertainty through von Neumann entropy~\citep{spectral_uncertainty}, and Generation Space Size uses EigenScore and its leave-one-out variant to characterize generation diversity~\citep{gss}. These studies establish the value of spectra for representation quality, uncertainty, and diversity, but do not use them to calibrate RLVR updates. To the best of our knowledge, SpeCal is the first to introduce hidden-state spectral calibration into unsupervised RLVR.

\section{Conclusion}
In this work, we presented SpeCal, a bi-level spectral calibration framework that improves the reliability of unsupervised RLVR without altering its underlying pseudo-reward. At the group level, SpeCalcombines spectral mass with an effective-rank band-pass gate to identify rollout groups exhibiting sufficient and well-organiz\linebreak[4]ed semantic variation, while at the rollout level, it evaluates each response against an effective-rank-adaptive leave-one-out subspace, thereby emphasizing trajectories supported by spectral directions independently established by the remaining rollouts. Experiments across Qwen- and Llama-based models demonstrate consistent improvements on both in-domain mathematical reasoning and out-of-domain general reasoning benchmarks. The compatibility experiments further show that SpeCal complements multiple unlabeled reward schemes, while the ablation results confirm that spectral spread, effective-rank calibration, and conditional rollout support provide complementary information. These findings suggest that hidden-state geometry offers a useful intermediate-granularity signal between coarse endpoint rewards and costly process supervision. Although the current evaluation focuses primarily on 8B-scale models and requires access to policy hidden states during training, future work could explore efficient spectral approximations, larger model scales, broader reasoning domains, and other forms of self-generated or weak supervision.
\bibliographystyle{ACM-Reference-Format}
\bibliography{references}

\appendix

\section{Detailed Theoretical Analysis}
\label{app:theory}

This appendix provides the assumptions and proofs for
Propositions~\ref{prop:endpoint_information_loss}
and~\ref{prop:marginal_spectral_consistency}.
The first result characterizes the information discarded by endpoint
summarization, while the second identifies the marginal geometric quantities
estimated by the two calibration levels of SpeCal.

\subsection{Proof of Endpoint Information Loss}
\label{app:endpoint_information_loss}

Recall that
$\mathcal{X}_q=\{(\hat a_j,\bm z_j)\}_{j=1}^{G}$
contains the decoded outcomes and response representations of a rollout group,
while $E_q=\Phi(\hat a_1,\ldots,\hat a_G)$ is a deterministic endpoint summary.
The reliability variable $C_q$ may represent, for example, whether the
pseudo-label is correct or whether the induced group-level supervision agrees with an oracle signal.

\begin{proof}[Proof of Proposition~\ref{prop:endpoint_information_loss}]
Since $E_q=\Phi(\mathcal{X}_q)$ is a deterministic function of
$\mathcal{X}_q$, conditioning on $\mathcal{X}_q$ fully determines $E_q$ and
therefore
\[
I(C_q;E_q\mid\mathcal{X}_q)=0.
\]
Applying the chain rule for mutual information in two orders gives
\[
I(C_q;\mathcal{X}_q,E_q)
=
I(C_q;\mathcal{X}_q)
=
I(C_q;E_q)+I(C_q;\mathcal{X}_q\mid E_q).
\]
Hence,
\[
I(C_q;\mathcal{X}_q)-I(C_q;E_q)
=
I(C_q;\mathcal{X}_q\mid E_q).
\]
By the conditional mutual-information identity,
\[
I(C_q;\mathcal{X}_q\mid E_q)
=
\mathbb{E}_{E_q,\mathcal{X}_q}
\left[
D_{\mathrm{KL}}
\left(
p(C_q\mid\mathcal{X}_q,E_q)
\,\Vert\,
p(C_q\mid E_q)
\right)
\right].
\]
Because $E_q$ is determined by $\mathcal{X}_q$, we have
$p(C_q\mid\mathcal{X}_q,E_q)=p(C_q\mid\mathcal{X}_q)$, yielding
Eq.~\eqref{eq:endpoint_information_loss}. Non-negativity follows from that of the KL divergence.

The gap is zero if and only if the endpoint summary $E_q$ is sufficient for
predicting $C_q$, namely,
$p(C_q\mid\mathcal{X}_q)=p(C_q\mid E_q)$, or equivalently,
$C_q\perp\!\!\!\perp\mathcal{X}_q\mid E_q$.
Conversely, if endpoint-equivalent rollout groups occur with positive
probability and two such groups $x_1$ and $x_2$ satisfy
$\Phi(x_1)=\Phi(x_2)$ but
$p(C_q\mid x_1)\neq p(C_q\mid x_2)$, then the endpoint summary cannot
distinguish their reliability, and the information gap is strictly positive.
\end{proof}

\subsection{Proof of Marginal Spectral Consistency}
\label{app:marginal_spectral_consistency}

\paragraph{Setup.}
Fix a question $q$ and a policy $\pi$. Let $\bm z_1,\bm z_2,\ldots\in\mathbb{R}^{D}$ be i.i.d.\ normalized response representations drawn from $P_\pi(\bm z\mid q)$, where $D=|\mathcal{L}|d$. Define the population mean and covariance as
\begin{equation}
\bm\mu_q=\mathbb{E}[\bm z\mid q], \qquad \mathbf{\Sigma}_q=\mathbb{E}\!\left[(\bm z-\bm\mu_q)(\bm z-\bm\mu_q)^\top\mid q\right].
\label{eq:app_population_covariance}
\end{equation}
For two independent draws $\bm z$ and $\bm z'$ from the same distribution, the marginal inter-rollout geometry is
\begin{equation}
\mathbf{\Sigma}^{\mathrm{marg}}_q=\frac{1}{2}\mathbb{E}\!\left[(\bm z-\bm z')(\bm z-\bm z')^\top\mid q\right]=\mathbf{\Sigma}_q.
\label{eq:app_marginal_equals_covariance}
\end{equation}
The equality follows by expanding the outer product and using $\mathbb{E}[\bm z\bm z'^\top\mid q]=\bm\mu_q\bm\mu_q^\top$. Thus, the marginal inter-rollout geometry coincides with the covariance of the prompt-conditional representation distribution.

Given $G$ sampled representations, let $\bar{\bm z}_G=G^{-1}\sum_{j=1}^{G}\bm z_j$, and let $\mathbf Z_c\in\mathbb{R}^{G\times D}$ stack $(\bm z_j-\bar{\bm z}_G)^\top$ row-wise. We define the centered empirical spectral matrix as
\begin{equation}
\widehat{\mathbf{\Sigma}}_{q,G}=\frac{1}{G-1}\mathbf Z_c^\top\mathbf Z_c.
\label{eq:app_empirical_covariance}
\end{equation}

\paragraph{Empirical and marginal calibration quantities.}
Let $\widehat{\lambda}_{G,1}\geq\cdots\geq\widehat{\lambda}_{G,D}\geq0$ be the eigenvalues of $\widehat{\mathbf{\Sigma}}_{q,G}$, including zero-valued directions when necessary. When the empirical spectral mass is nonzero, define
\begin{equation}
\begin{aligned}
\widehat m_G&=\frac{G-1}{G}\operatorname{tr}(\widehat{\mathbf{\Sigma}}_{q,G}), \qquad
\widehat p_{G,k}=\frac{\widehat\lambda_{G,k}}{\sum_{\ell=1}^{D}\widehat\lambda_{G,\ell}},\\
\widehat{\operatorname{ER}}_G&=\exp\!\left(-\sum_{k=1}^{D}\widehat p_{G,k}\log\widehat p_{G,k}\right), \qquad
\widehat\eta_G=\frac{\widehat{\operatorname{ER}}_G}{r_G},
\end{aligned}
\label{eq:app_empirical_spectral_statistics}
\end{equation}
where $r_G=\min(G-1,D)$ and $0\log0$ is defined as zero. The finite-group calibration coefficient is then
\begin{equation}
\widehat w_G(q)=\widehat m_G\exp\!\left(-\frac{(\widehat\eta_G-\mu_G)^2}{2\tau_G^2}\right).
\label{eq:app_empirical_group_weight}
\end{equation}
This is the empirical quantity denoted by $w_G(q)$ in the main text. For a spectrally collapsed group with zero empirical mass, we set $\widehat w_G(q)=0$.

Let $\lambda_{q,1}\geq\cdots\geq\lambda_{q,D}\geq0$ denote the eigenvalues of $\mathbf{\Sigma}^{\mathrm{marg}}_q$. We define the corresponding marginal quantities by
\begin{equation}
\begin{aligned}
m_q^\star&=\operatorname{tr}(\mathbf{\Sigma}^{\mathrm{marg}}_q), \qquad
p_{q,k}^\star=\frac{\lambda_{q,k}}{\sum_{\ell=1}^{D}\lambda_{q,\ell}},\\
\operatorname{ER}_q^\star&=\exp\!\left(-\sum_{k=1}^{D}p_{q,k}^\star\log p_{q,k}^\star\right), \qquad
\eta_q^\star=\frac{\operatorname{ER}_q^\star}{D}.
\end{aligned}
\label{eq:app_marginal_spectral_statistics}
\end{equation}
The marginal group-level target is
\begin{equation}
w_G^\star(q)=m_q^\star\exp\!\left(-\frac{(\eta_q^\star-\mu_G)^2}{2\tau_G^2}\right).
\label{eq:app_population_group_weight}
\end{equation}

For rollout-level calibration, let $k_q^\star=\lceil\operatorname{ER}_q^\star\rceil$, and let $\mathbf V_q^{(k_q^\star)}$ contain the corresponding top eigenvectors of $\mathbf{\Sigma}^{\mathrm{marg}}_q$. The marginal projector and conditional support are
\begin{equation}
\mathbf P_q^\star=\mathbf V_q^{(k_q^\star)}\mathbf V_q^{(k_q^\star)\top}, \qquad
s_R^\star(\bm z)=\frac{(\bm z-\bm\mu_q)^\top\mathbf P_q^\star(\bm z-\bm\mu_q)}{\|\bm z-\bm\mu_q\|_2^2}.
\label{eq:app_population_rollout_support}
\end{equation}
For a finite group, $\widehat s_{R,j}$ denotes the support computed by the same leave-one-out construction as in the main method, using the top $\widehat k_G=\lceil\widehat{\operatorname{ER}}_G\rceil$ available directions.

\begin{assumption}[Spectral regularity]
\label{ass:spectral_regularity}
For the fixed question $q$ and policy $\pi$, the following conditions hold:
\begin{enumerate}[label=(A\arabic*),leftmargin=*,itemsep=2pt,topsep=3pt]
    \item The representations are i.i.d.\ and satisfy $\mathbb{E}[\|\bm z\|_2^2\mid q]<\infty$, with $\operatorname{tr}(\mathbf{\Sigma}^{\mathrm{marg}}_q)>0$.
    \item The marginal effective rank is separated from integer boundaries, i.e., $\operatorname{dist}(\operatorname{ER}_q^\star,\mathbb{Z})>0$, so the retained dimension $k_q^\star$ is locally stable.
    \item The marginal covariance has a positive eigengap at $k_q^\star$, namely $\lambda_{q,k_q^\star}>\lambda_{q,k_q^\star+1}$, and $\|\bm z_j-\bm\mu_q\|_2>0$ with probability one. The eigengap condition is unnecessary when $k_q^\star=D$.
\end{enumerate}
\end{assumption}

\begin{proof}[Proof of Proposition~\ref{prop:marginal_spectral_consistency}]
We first establish the finite-sample unbiasedness of the empirical geometry. The pairwise-scatter identity gives
\begin{equation}
\widehat{\mathbf{\Sigma}}_{q,G}=\frac{1}{G(G-1)}\sum_{1\leq i<j\leq G}(\bm z_i-\bm z_j)(\bm z_i-\bm z_j)^\top.
\label{eq:app_pairwise_scatter}
\end{equation}
Each pair $(\bm z_i,\bm z_j)$ has the same distribution as the independent pair $(\bm z,\bm z')$, and the sum contains $G(G-1)/2$ terms. Therefore,
\begin{equation}
\mathbb{E}[\widehat{\mathbf{\Sigma}}_{q,G}\mid q]=\frac{1}{2}\mathbb{E}\!\left[(\bm z-\bm z')(\bm z-\bm z')^\top\mid q\right]=\mathbf{\Sigma}^{\mathrm{marg}}_q.
\label{eq:app_empirical_unbiasedness}
\end{equation}

We next consider the large-group limit. By the strong law of large numbers,
\begin{equation}
\bar{\bm z}_G\xrightarrow[G\to\infty]{\mathrm{a.s.}}\bm\mu_q, \qquad
\frac{1}{G}\sum_{j=1}^{G}\bm z_j\bm z_j^\top\xrightarrow[G\to\infty]{\mathrm{a.s.}}\mathbb{E}[\bm z\bm z^\top\mid q].
\label{eq:app_moment_convergence}
\end{equation}
It follows that
\begin{equation}
\widehat{\mathbf{\Sigma}}_{q,G}\xrightarrow[G\to\infty]{\mathrm{a.s.}}\mathbf{\Sigma}^{\mathrm{marg}}_q.
\label{eq:app_covariance_convergence}
\end{equation}
Because $D$ is fixed, the convergence also holds in operator norm. The ordered empirical eigenvalues therefore converge to the corresponding marginal eigenvalues. For all sufficiently large $G$, we also have $r_G=D$. Under Assumption~\ref{ass:spectral_regularity}, spectral mass, entropy effective rank, normalized effective rank, and the Gaussian gate are continuous at the marginal geometry. The continuous-mapping theorem consequently yields
\begin{equation}
\widehat w_G(q)\xrightarrow[G\to\infty]{\mathrm{a.s.}}w_G^\star(q).
\label{eq:app_group_weight_convergence}
\end{equation}

For rollout-level calibration, removing a fixed observation does not change the limiting mean or covariance. Hence, the leave-one-out spectral matrix converges to $\mathbf{\Sigma}^{\mathrm{marg}}_q$. Assumption~\ref{ass:spectral_regularity} ensures that $\widehat k_G=k_q^\star$ for all sufficiently large $G$ and that the retained eigenspace is uniquely defined. Standard eigenspace perturbation results then imply
\begin{equation}
\widehat{\mathbf P}_{-j,G}^{(\widehat k_G)}\xrightarrow[G\to\infty]{\mathrm{a.s.}}\mathbf P_q^\star.
\label{eq:app_projector_convergence}
\end{equation}
Together with the convergence of the leave-one-out mean, applying the continuous-mapping theorem to the normalized quadratic form gives
\begin{equation}
\widehat s_{R,j}\xrightarrow[G\to\infty]{\mathrm{a.s.}}s_R^\star(\bm z_j).
\label{eq:app_support_convergence}
\end{equation}
Equations~\eqref{eq:app_group_weight_convergence} and~\eqref{eq:app_support_convergence} complete the proof.
\end{proof}

\paragraph{Connection to the Gram matrix used by SpeCal.}
The method computes $\mathbf K_G=\mathbf Z_c\mathbf Z_c^\top$, whereas the analysis uses $\widehat{\mathbf{\Sigma}}_{q,G}=(G-1)^{-1}\mathbf Z_c^\top\mathbf Z_c$. These matrices share the same nonzero singular spectrum up to the deterministic factor $G-1$. Therefore, they induce the same normalized eigenspectrum and entropy effective rank, while their spectral masses satisfy
\begin{equation}
m_G=\frac{1}{G}\operatorname{tr}(\mathbf K_G)=\frac{G-1}{G}\operatorname{tr}(\widehat{\mathbf{\Sigma}}_{q,G}).
\label{eq:app_gram_covariance_relation}
\end{equation}
The consistency result thus applies directly to the spectral quantities used in SpeCal.

\paragraph{Scope of the result.}
Proposition~\ref{prop:marginal_spectral_consistency} distinguishes finite-sample unbiasedness from nonlinear plug-in consistency. The empirical geometry matrix is an unbiased estimator of the marginal inter-rollout geometry, whereas the group weight and conditional support converge to their marginal targets as the rollout group grows. The proposition does not claim that these nonlinear quantities are unbiased at finite $G$, nor that marginal spectral geometry is identical to ground-truth correctness.

\section{Additional Experimental Setups}
\label{app:exp_setup}

We limit input prompts to \(1{,}024\) tokens and generated responses to
\(4{,}096\) tokens to accommodate long-form reasoning. For rollout
generation, we use the following system prompt:
``\texttt{Let's think step by step and output the final answer within
\textbackslash boxed\{\}.}''
Rollouts are generated using
\texttt{vLLM}\footnote{\url{https://github.com/vllm-project/vllm}}
with \(G=8\) responses per prompt, a temperature of \(1.0\), and
top-\(p\) of \(1.0\). By default, all models and baselines are trained for a total of 500 global steps.

We employ \texttt{Math-Verify}\footnote{\url{https://github.com/huggingface/Math-Verify}} as the sole outcome reward function, deliberately avoiding auxiliary rewards for formatting constraints or response lengths to prevent reward hacking. Finally, rollout generation is accelerated via the vLLM framework2 with 8 rollouts per prompt, using a fixed temperature of 1.0 and a top-p value of 1.0.
For hidden-state extraction, we select eight layers evenly spaced across the network depth for every model backbone by default. Additional ablations on this layer-selection strategy are provided in Appendix~\ref{app:layer}.

\section{Additional Experimental Results}
\label{app:additional_results}
\subsection{Effect of Representation Construction}
\label{app:layer}
\begin{figure*}[t]
    \centering
    \includegraphics[width=0.98\textwidth]
    {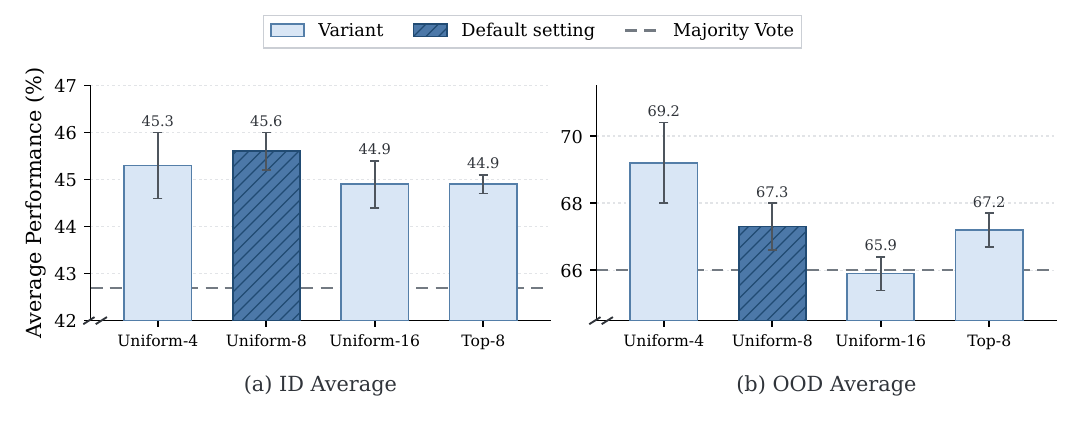}
    \caption{
    Comparison of different hidden-state representation constructions on
    ID and OOD evaluations. Uniform-$k$ selects $k$ layers uniformly across
    the model, whereas Top-8 selects the top eight layers. All variants
    otherwise use layer concatenation and response-token mean pooling.
    The dark hatched bar marks the default Uniform-8 configuration, while
    the gray dashed line denotes the Majority Vote baseline. Values above
    the bars report average performance. Panel-specific y-axis ranges with
    explicit breaks are used for improved readability.
    }
    \label{fig:representation_construction}
\end{figure*}

To investigate how layer selection affects the hidden-state representation used by SpeCal, we compare four construction strategies under otherwise identical training and evaluation settings. Uniform-$k$ evenly selects $k$ layers across the model depth, whereas Top-8 uses only the top eight layers. For all variants, the selected layer representations are concatenated, followed by mean pooling over the response tokens. As shown in Fig.~\ref{fig:representation_construction}, all four variants outperform the Majority Vote baseline of 42.7 on the ID evaluation. Uniform-8 achieves the best ID average of 45.6, followed closely by Uniform-4 at 45.3, while Uniform-16 and Top-8 both obtain 44.9. At the same layer count, Uniform-8 exceeds Top-8 by 0.7 points on ID and 0.1 points on OOD, suggesting that covering different model depths is preferable to relying exclusively on upper-layer features. The OOD results exhibit a different trend: Uniform-4 achieves the highest average of 69.2, outperforming the OOD baseline by 2.2 points, whereas Uniform-8 and Top-8 provide smaller improvements of 0.3 and 0.2 points, respectively. Increasing the number of uniformly selected layers to 16 reduces the OOD average to 65.9. Overall, these results show that incorporating more hidden states does not monotonically improve performance; a compact selection distributed across the model depth can retain sufficient task information while avoiding potentially redundant representations.

\subsection{Training Efficiency}
We compare the end-to-end training efficiency of different methods by measuring their average wall-clock time per optimization step under the same hardware and training configuration. As shown in Table~\ref{tab:runtime_comparison}, SpeCal requires 198 seconds per step on Qwen3-8B-Base and 251 seconds on DeepSeek-R1-Distill-Llama-8B. Compared with TTRL, SpeCal introduces only a 3.1\% runtime overhead on Qwen3-8B-Base and is numerically 1.2\% faster on DeepSeek-R1-Distill-Llama-8B, indicating that the additional spectral calibration incurs limited computational overhead. SpeCal is also substantially more efficient than Co-rewarding, reducing the per-step training time by 19.2\% and 41.2\% on the two base models, respectively. SCRL remains the fastest method, particularly on Qwen3-8B-Base, where SpeCal requires 76.8\% more time; however, this gap decreases to only 2.0\% on DeepSeek-R1-Distill-Llama-8B. Overall, although SpeCal is not the fastest method in every setting, it maintains runtime comparable to TTRL and avoids the considerably larger overhead of Co-rewarding across both model families.

\begin{table}[t]
    \centering
    \caption{
    Average training time per optimization step under two
    base models. All methods are evaluated using the same hardware and
    training configuration.
    }
    \label{tab:runtime_comparison}
    \begin{tabular}{lcc}
        \toprule
        & \multicolumn{2}{c}{Time per Step (s) $\downarrow$} \\
        \cmidrule(lr){2-3}
        Method
        & Qwen3-8B-Base
        & DeepSeek-R1-Distill-Llama-8B \\
        \midrule
        TTRL      & 192 & 254 \\
        SCRL      & 112 & 246 \\
        Co-rewarding     & 245 & 427\\
        \midrule
        \rowcolor{Gray}\textbf{SpeCal} & 198 & 251 \\
        \bottomrule
    \end{tabular}
\end{table}

\section{Pseudocode}
The complete training procedure of SpeCal is summarized in
Algorithm~\ref{alg:specal}.

\begin{algorithm*}[!t]
\caption{SpeCal}
\label{alg:specal}
\KwIn{
Unlabeled dataset $\mathcal{D}_{\mathrm U}$,
policy $\pi_\theta$,
unlabeled reward scheme $\mathcal{R}_{\mathrm U}$,
training steps $T$,
rollout size $G$,
selected layers $\mathcal{L}$,
calibration parameters
$\{\mu_G,\tau_G,w_R^{-},w_R^{+}\}$
}
\KwOut{Optimized policy $\pi_\theta$}

\BlankLine
\For{\textnormal{step } $t=1$ \KwTo $T$}{
    Sample an unlabeled mini-batch
    $\mathcal{B}\sim\mathcal{D}_{\mathrm U}$\;

    \BlankLine
    \ForEach{$q_i\in\mathcal{B}$}{
        Generate $G$ rollouts
        $\{y_{i,j}\}_{j=1}^{G}
        \sim\pi_\theta(\cdot\mid q_i)$\;
        
        Construct unlabeled rewards using $\mathcal{R}_{\mathrm U}$
        and compute the original advantages
        $\{\widehat A^{\mathrm U}_{i,j}\}_{j=1}^{G}$\;

        \BlankLine
        \tcp{\textbf{Group-Level Spectral Calibration}}
        Extract response representations and construct the centered
        group spectrum via
        Eqs.~\eqref{eq:cross_layer_response_repr}--%
        \eqref{eq:group_gram_matrix}\;
        
        Compute the group weight $w_G(q_i)$ and adaptive spectral
        bandwidth $k_G$ via
        Eqs.~\eqref{eq:group_effective_rank}
        and~\eqref{eq:group_er_gated_mass}\;

        \BlankLine
        \tcp{\textbf{Rollout-Level Spectral Calibration}}
        \ForEach{$y_{i,j}$ in the rollout group}{
            Construct the effective-rank-truncated leave-one-out
            subspace via
            Eqs.~\eqref{eq:loo_svd}--%
            \eqref{eq:loo_topk_projector}\;
            
            Compute the rollout weight $w_{R,i,j}$ via
            Eqs.~\eqref{eq:conditional_spectral_support}
            and~\eqref{eq:trajectory_weight}\;
        }

        Calibrate the original advantages using $w_G(q_i)$ and
        $\{w_{R,i,j}\}_{j=1}^{G}$ via
        Eq.~\eqref{eq:SpeCal_advantage}\;
    }

    Update $\pi_\theta$ with the calibrated GRPO objective in
    Eq.~\eqref{eq:SpeCal_objective}\;
}

\KwRet{Optimized policy $\pi_\theta$}\;
\end{algorithm*}

\end{document}